# Nighttime Driver Behavior Prediction Using Taillight Signal Recognition via CNN-SVM Classifier

Amir Hossein Barshooi and Elmira Bagheri

*Abstract*—This paper aims to enhance the ability to predict nighttime driving behavior by identifying taillights of both human-driven and autonomous vehicles. The proposed model incorporates a customized detector designed to accurately detect front-vehicle taillights on the road. At the beginning of the detector, a learnable pre-processing block is implemented, which extracts deep features from input images and calculates the data rarity for each feature. In the next step, drawing inspiration from soft attention, a weighted binary mask is designed that guides the model to focus more on predetermined regions. This research utilizes Convolutional Neural Networks (CNNs) to extract distinguishing characteristics from these areas, then reduces dimensions using Principal Component Analysis (PCA). Finally, the Support Vector Machine (SVM) is used to predict the behavior of the vehicles. To train and evaluate the model, a large-scale dataset is collected from two types of dash-cams and Insta360 cameras from the rear view of Ford Motor Company vehicles. This dataset includes over 12k frames captured during both daytime and nighttime hours. To address the limited nighttime data, a unique pixel-wise image processing technique is implemented to convert daytime images into realistic night images. The findings from the experiments demonstrate that the proposed methodology can accurately categorize vehicle behavior with 92.14% accuracy, 97.38% specificity, 92.09% sensitivity, 92.10% F1-measure, and 0.895 Cohen's Kappa Statistic. Further details are available at https://github.com/DeepCar/Taillight_Recognition.

## I. INTRODUCTION

NIGHT driving is much more difficult than during the day and requires more skill and caution. The visibility of the drivers in the dark has dropped significantly and it will be between 110 and 150 meters in the best case [1]. In the past years, Autonomous Vehicles (AVs) have shown relatively good performance in the dark, one of the most prominent ones is the Ford Fusion vehicle [2]. Ford Motor Company tested this vehicle in a condition where all its lights were off and the road did not have any lighting system. The darkness of the test route was so much that it was not possible

The authors are with the School of Automotive Engineering, Iran University of Science and Technology, (e-mail: a_barshooi@alumni.iust.ac.ir, elmira_bagheri@alumni.iust.ac.ir). All Authors have equal contributions.

(Corresponding author: Amir Hossein Barshooi)

to see the vehicle itself with the naked eye and without night vision cameras. Despite all the efforts made, AVs are still far from the ideal point. For example, the speed of these vehicles is limited to 60 miles per hour, and adverse weather conditions are not taken into account [2].

By taking control of the vehicle in complex environmental conditions, AVs play a crucial role in enhancing safety and decreasing road accidents [3]. A better understanding of the environment, less reaction delay, and greater dominance over vehicle components have made these vehicles far better decisions than humans in critical situations. Much of the awareness of AVs with any level of automation is achieved by machine learning techniques [4]. Today, the evolution of autonomous, connected, shared, and environmentally friendly vehicle technology has transformed the Intelligent Transportation System (ITS) more than in the past. In addition, by reducing human interaction in driving, it has been able to achieve other secondary goals, such as safety and traffic management [5].

Collecting information is the most basic step in AVs, through which decisions are made. In these vehicles, information from the surrounding environment is collected with a range of sensors and aftermarket devices. The characteristics and limitations of each, cause different operational design domains. Designers of AVs face the challenge of establishing a tradeoff between accuracy, field of view, system complexity, and cost. They accomplish this by carefully selecting suitable sensors, determining their installation positions, and synchronizing them with each other. In spite of the researchers' endeavors to find solutions to improve the performance of AVs in different challenging conditions, there is still a long way from the ideal.

By examining the studies, it is clear that most AV accidents either happen at night and in the dark, or inappropriate weather conditions are somehow involved in them [6]. By interacting with the surrounding environment, the cameras in AVs calculate the position of the vehicles at both the local and global levels and display the route from the road lines [3]. Raindrops, snowflakes, and dust can easily blind the vision of these cameras by forming a thin layer on the cameras and covering the white lines of the road. Meanwhile, other equipment such as LiDAR and sound sensors cannot be completely certified, because this condition will harm their performance. For example, snowflakes play a similar role as a shield by absorbing a large portion of sound signals. Also, the darkness of the air will disrupt the performance of vehicles by reducing the visibility of the environment and showing the road lines faintly [7]. To solve the challenges caused by light, equipment such as infrared cameras or road signs with the ability to reflect light can be used. However, experience has shown that the utilization of infrared cameras is not computationally efficient for real-world applications. Additionally, the functionality of reflective road signs is rendered ineffective in unfavorable weather conditions.

This study presents an innovative technique that integrates image processing and deep learning to enhance the performance of AVs and human-driven vehicles in low light or adverse weather conditions. The method requires minimal facilities and can be implemented on any vehicle. The proposed method predicts the behavior of front drivers from the signals of the taillights and informs other vehicles based on the wireless communication network. In general, the innovations of this study can be described as follows:

- Collecting large-scale datasets from the rear view of vehicles along with labeling Regions of Interest (ROIs) of taillights and brake lights;
- Introducing a novel pixel-wise image-to-image translation framework to create nighttime images from daytime along with modeling different weather conditions;
- Designing a weighted binary mask to guide the model in finding rich features;
- Extracting features with different depths from images with the help of Convolutional Neural Networks (CNNs) and classification with Support Vector Machines (SVM);
- Using PCA to reduce the dimensions of images and improve accuracy of classification.

The rest of this study is organized as follows: Section II focuses on examining AV research in the context of the Internet of Things (IoT), cloud computing, and deep learning. Section III introduces the collected dataset. Details about the proposed approach and research objectives can be found in section IV. The outcomes are presented in section V. Finally, conclusions and future works for research are discussed in section VI.

## II. Literature Review

By examining the driver-assist features in increasing the safety of connected vehicles and self-driving vehicles in nighttime driving, it is possible to study the activities carried out in the two fields of IoT and machine learning.

### A. Internet of Things and blockchains

The Internet of Things (IoT), blockchains, and cloud computing [8] are three key elements in AVs. Kong et al. [9], designed an IoT-cloud-based system that can deal with adverse weather conditions by sharing data obtained from various sensors of AVs. In addition, incorporating cloud computing into this system enhanced its performance in accurately detecting and recognizing small objects present on the road. Ravikumar and Kavitha [10], combined the neuro-fuzzy inference system with the black window algorithm and introduced a new AV driver scheme in the context of the IoT. In 2018, Elshaer et al. [11], fabricated an AV based on the IoT. The hardware part of this vehicle consisted of five ECUs, a CAN bus, and various sensors. It was able to obstacle collision avoidance, and recognize traffic lights. In this plan, users in the monument area registered their tour requests through a unique user interface and waited for the vehicle to arrive. The designed AVs found the location of the users through Google Maps, the IoT, and GPS, and by moving on a planned route, it took them to the destination point.

Ford Motor Company, a leading global automaker, has been implementing blockchain technology in the automotive sector for a number of years, in collaboration with three other companies: BMW, General Motors, and Renault. The consortium formed the Mobility Open Blockchain Initiative (MOBI) to focus on securing, controlling, and managing information related to the automotive industry with blockchain technology [12]. In this regard, Rathee et al. [13], presented a security mechanism for AVs using blockchain technology. In another study, Kamble et al. [14], investigated common applications of blockchain in AVs such as charging stations, parking, insurance, and vehicle lifecycle in addition to the security mechanism, and also reviewed the state-of-the-art work done in each.

### B. Computer vision and deep learning

Pirhonen et al. [15], presented a new hybrid method for the detection of brake light of vehicles from a distance of 150 meters, using YOLOv3 and a random forest classifier. They evaluated their algorithm in the city of Helsinki in southern Finland under different conditions and achieved an accuracy of 81.8%. Chen et al. [16], identified brake lights of vehicles in urban areas and highways with 79% accuracy using Nakagami-m distribution. Tong et al. [17], proposed a real-time strategy to detect vehicle taillights by combining a YOLOv4 detector with a Feature Pyramid Network (FPN) based module. Also, they evaluated their system under various weather conditions on different roads by collecting a dataset consisting of 3316 images from the BDD100K collection. Their proposed system could improve the mean Average Precision (mAP) criterion by 24.63%. Kavya et al. [18], determined vehicles at night based on the red color of the vehicle's taillights and applied morphological operations.

Vu et al. [19], proposed a new robust algorithm to improve the performance of surveillance systems in detecting and predicting the type of vehicles in night-time driving. The method is designed by analyzing the driving patterns from the headlights of vehicles. It is also classified into two classes such as two-wheelers, and four-wheelers in disorderly traffic. This method is evaluated on two DBP02 and VVK01 datasets and was able to reach accuracies of 79.46 and 82.91 percent respectively. Recently, Wu et al. [20], were able to improve the speed, accuracy, and calculations of vehicle detectors in the dark by embedding the Ghostnet V2 module in YOLO7. The results showed that this method could achieve a speed of 47 frames per second while reducing floating-point operations by 60%.

## III. INTRODUCING DATASET

One of the primary requirements in the proposed method is to collect data from the roads on traffic conditions, which includes the taillights of the vehicles. In this paper, data is collected from the rear angles of the vehicles on the road and labeled the area where the left, right, and brake lights are located in YOLO format. This dataset includes a total of 110 videos and 3800 images. Each vehicle is in one of the states of "braking", "running", "left-turn", and "right-turn". The dataset is collected from the web and social media. The length of each video is 12 seconds on average. This dataset has various lighting conditions and qualities, more than 70% are taken using an Insta360 camera, and the rest with normal cameras or dash-cams. Insta360 is a spherical-shaped professional virtual reality camera that can record 360-degree 3D videos with a maximum resolution of 8K with the help of 6 separate cameras and lenses [21]. By installing this camera with an invisible monopod on the trunk of vehicles, the signals of the taillights of vehicles can be recorded. The vehicles in this dataset belong to Ford Motor Company and include various ranges of this company's products from 2008 to 2023. As it is considered, there is no similarity for this dataset in the literature. Among the most famous models in this dataset, are Mustang, Fusion, Raptor, Shelby, and Explorer.

Although the proposed method is based on Ford Motor Company products, it can be generalized to other vehicles around the world for two reasons. First, the vehicle models selected in this paper include different body styles from sedans to SUVs in different years. Secondly, colors and patterns were the same in all of the vehicles, however, had different sizes and shapes taillights.

At a wavelength of 700 nm, the red spectrum exhibits the least frequency among light waves and has traditionally been associated with the symbol of caution and alertness. The human eye is less sensitive to this light compared to other lights. Also, due to the low effect of air molecules in its dispersion, it can travel the greatest distance in snowy, rainy, or dusty conditions [22]. Fig. 1. illustrates an instance of the lighting impact of vehicles in Ford Motor Company products across four classes, namely "braking," "running," "left-turn," and "right-turn".

## IV. EXPERIMENTAL DETAILS

In today's vehicles, if drivers want to change lanes, turn signals are activated automatically and warn the drivers behind. This is also true about AVs. The method presented in this study can be easily implemented with the help of a mobile camera or dash-cam on all normal vehicles and AVs with different levels. The images taken by these cameras are sent to a cloud server. From the taillight of the front vehicles, their behavior is classified into four classes braking, running, left-turn, and right-turn. Also, it is transmitted to other drivers (or AVs) with the help of the IoT. This notification gives the drivers a general view of the traffic flow and based on that, choose the most suitable lane to continue their movement. For example, drivers find out that out of 100 vehicles moving in all available lanes at a distance of 500 meters, 10 vehicles intend to change lanes and move to the right. As a result, the volume of traffic in the right lanes will soon increase, and drivers (or AVs) should switch to the left lanes as soon as possible. At the beginning of this design, a detector is installed to draw the smallest possible bounding box in which all three left, right and brake lights are located. Then, the images are fed as input to the CNN with VGGNet architecture. The last fully connected layer's results are put through the PCA algorithm to decrease its dimensions and only the significant variables are taken from it.

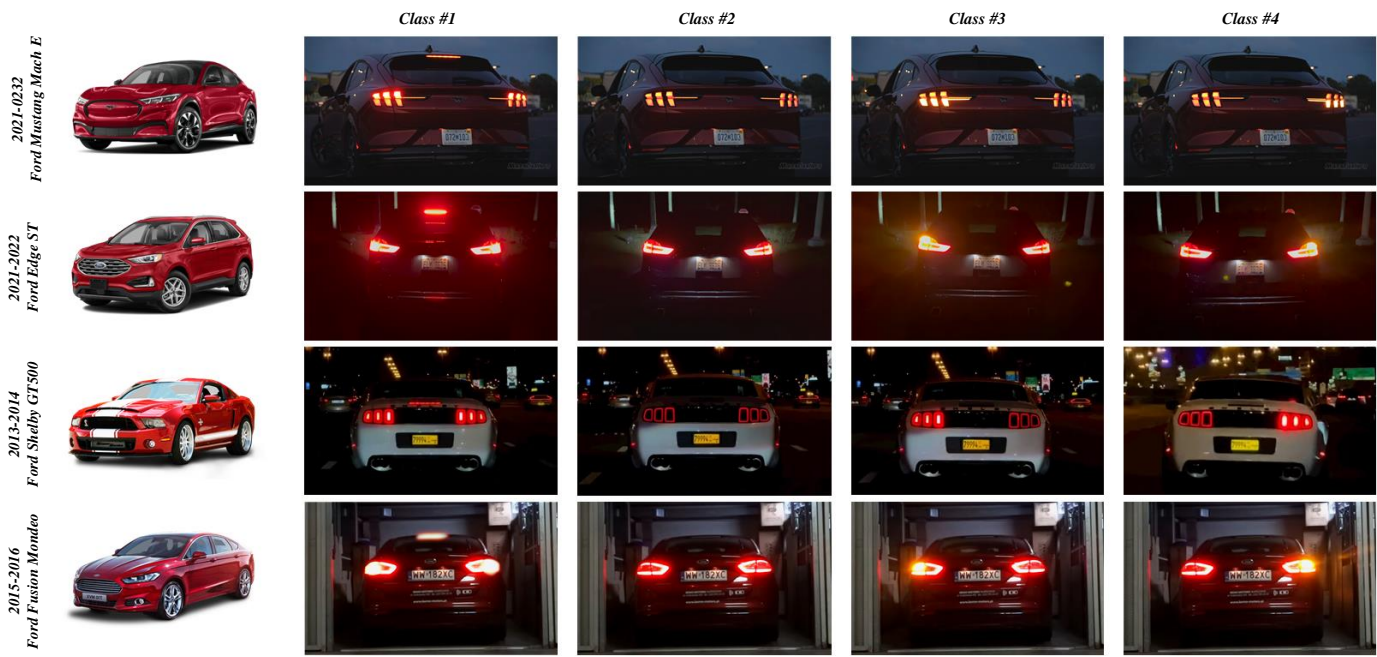

**Fig. 1.** An example of lightening effect of the Ford Motor Company products in four classes; #1) braking, #2) running, #3) left-turn, and #4) right-turn

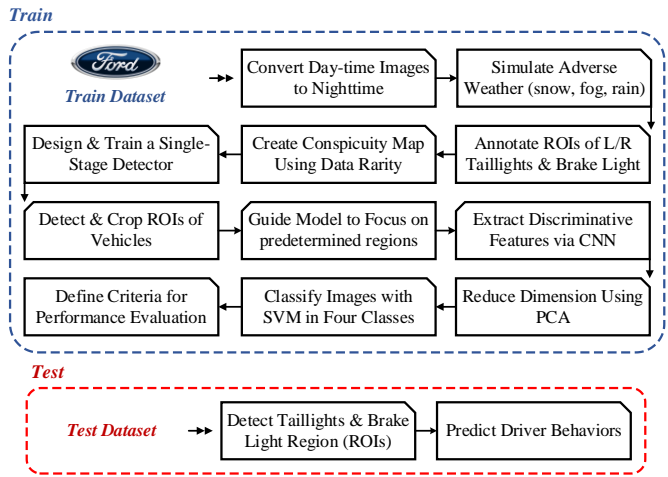

**Fig. 2.** Flow chart of the proposed design for Ford Motor Company vehicles

Finally, with the help of the SVM classifier embedded at the end of the model, the class of each image is predicted. The details of the design are shown in Fig. 2.

### A. Image-to-image translation

Image-to-image translation is a prevalent technique used in computer vision and image processing tasks, aimed at transforming the source image to a target image of choice [23]. Today, this technique plays a special role in various applications such as image enhancement, style transfer [24], season transfer, and image restoration [25]. In this section, inspired by image-to-image translations, a new pixel-wise framework is designed to transfer images from daytime hours to nighttime. This method has a concept close to the season transfer algorithm in image-to-image translations, where instead of changing the seasons, the hours of the day are changed. Next, to simulate different weather models, image corruption blocks are implemented on the images to transfer them from the normal domain to the adverse weather domain.

#### 1) Synthetic nighttime image generation:

The main focus of this study is to predict the behavior of vehicles in difficult conditions, especially in the hours of darkness. The data format available on the web is taken in daylight and it is rare to find images from the rear angle of vehicles at night. In this section, considering image processing techniques, this limitation is resolved. Applying transformations on the pixel values of each image, changed from daytime to nighttime. The mentioned algorithm is performed in two steps as follows.

In the first step, the images are sequentially retrieved and the contour that encloses the vehicles is identified as the foreground, while the remaining pixels that represent the road are classified as the background. In the second step, the parameters related to color, light, white balance, and gamma of both foreground and background pixels are changed separately. The parameters of contrast, brightness, exposure, highlights, and shadows, which are a subset of color, are chosen for the foreground equal to 29, -56, 44, 18, and -55 percent of their initial values, respectively. Although, these values for the background are equal to 31, -71, -37, -72, and -5 percent, respectively. The Hue values [26] as one of the color parameters are set to 0 in both the foreground and background. The amount of saturation, as another parameter, for the foreground and background, changed to 13 and 18 percent of the initial value, respectively. Temperature and tint parameters form the characteristics of white balance in images. In the foreground, these values are selected as 6 and 8 percent of initial values, and in the background, -56 and 3 percent of the initial values, respectively. A combination of red, green, and blue channels, known as gamma, are selected in the foreground at 10, 1, and 1 percent of the current value in each image. In the background, these values are considered equal to 10, 4, and -23 percent. The steps of transferring daytime images to nighttime images are shown in Fig. 3a. In addition, examples of real images collected at night with dash-cams and Insta360 cameras can be seen in Fig. 3b.

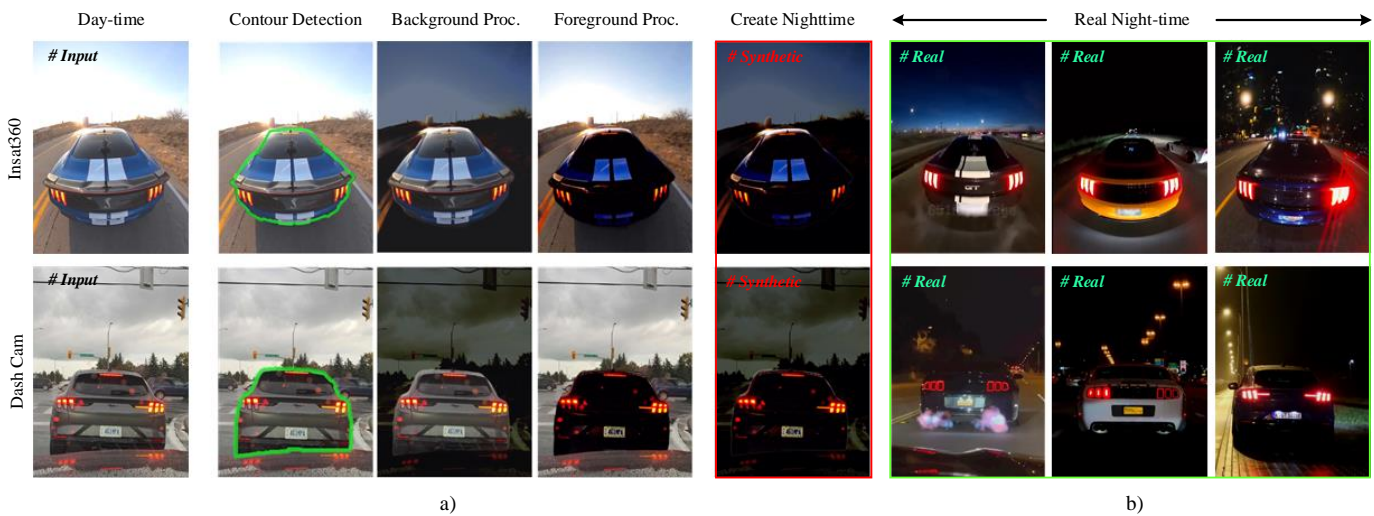

a)                                                                                              b)

**Fig. 3.** a) the steps of creating synthetic nighttime images from daytime images in both types of Insta360 cameras and dash-cams, b) an example of real images taken at night

### 2) Adverse weather model generation:

For data-driven methods to perform well in the real world and challenging conditions, they must be exposed to various environments during the training process [27]. Unfavorable weather, low light conditions, low resolution, dust on lenses, and camera vibration are among these conditions. Since collecting the required data in a wide range of these challenges is very time-consuming and costly, in this paper, an innovative method is used to create images in different conditions. The image corruption block [28] designed in this section consisted of 13 image masks in three different levels: mild, moderate, and severe. Seven of these masks are chosen for the train-set and the reminder for test-set. Rain Blur, Snow, Fog, Alpha Blend, Frosted Glass Blur, Lens Defect, and JPEG masks are applied to the images during the training phase. These masks are used to enhance the performance of the model. Zoom Blur, Frost, Contrast, Rain Drop, Shot Noise, and Pixelate masks served as benchmarks to assess the model's performance. Each of the challenging environmental conditions and unfavorable weather with one of these masks has been modeled. In other words, JPEG and Pixelate masks are equivalent to low camera resolution, and Fog and Contrast masks are equivalent to thick fog and smoke. Snow, Shot Noise, and Frost masks are equivalent to snowy weather, dirt, or freezing of the lens, respectively. Rain Drop, and Rain Blur make a rain effect. The Frosted Glass Blur, and Alpha Blend masks are also designed to simulate camera shake caused by car movement on uneven roads. In general, the design goals of these masks can be examined from three aspects, which are proven in section V. First, by combining all masks with conventional data augmentation techniques and inputting them into deep learning-based networks, the model's robustness in demanding conditions is enhanced. Secondly, utilizing smartphone cameras enhances the model's ability to adapt for implementation in human-driven vehicles, thus increasing its generalization. Thirdly, applying these masks to the dataset is also seen as a form of data augmentation, which helps prevent the model from overfitting during training [27]. Examples of mapping some of these masks in severe level on the training and testing data are shown in Fig. 4.

### B. Blended ROI detection

Driving for a long time, especially at night, makes it difficult for drivers to determine the distance and predict the movements of front vehicles. As a result, it is crucial for drivers following behind to pay attention to brake lights and turn signals. In this paper, similar to the behavior of drivers, the signal of the right and left taillights along with the brake light are considered effective ROIs in predicting the vehicle's path. Determining these ROIs, similar to other detection and tracking tasks under ambient light conditions, is one of the most challenging topics in computer vision, which has been receiving a lot of focus lately. Meanwhile, adverse weather conditions aggravate these tasks.

This section introduces a model that can rapidly and precisely identify the designated ROIs and subsequently transfer them to the next stages for feature extraction and selection. This model is based on the You Only Learn One Representation (YOLOR) algorithm [29], which is extensively utilized in real-time object detection assignments. The class of objects is determined by the exchange of explicit and implicit knowledge.

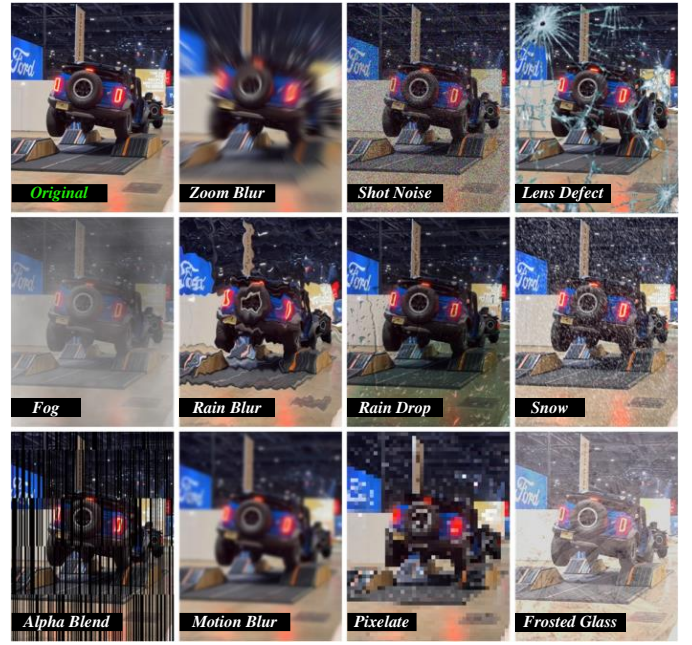

**Fig. 4.** Examples of applying real-world environment modeling masks on images

Before feeding the images to this algorithm, learnable pre-processing blocks are designed. Customized VGG-19 encoder assists in extracting deep features from the input images. To obtain a wide range of deep features, it is better to train the network on a large-scale dataset with different objects. In this issue, the ImageNet dataset is selected. Similar to [30], the fully connected and pooling layer from the VGG-19 network is removed. Then, in each layer and for each feature, the data rarity is calculated [31]. To be more precise, there are $2\times64$ features at level 1, and the data rarity is calculated for each. These 64 outputs are fused with certain weights and formed the "conspicuity map".

Then, the dataset collected in section III is fed one by one to this model, and the conspicuity map of each one is obtained. Designing conspicuity maps, instead of using handcrafted pre-processing filters, is one of the ideas presented in this paper. The edges of the main components of the car, such as the rear window, the trunk, and the contour of the car, are visible in these maps. In the next step, the original raw images are merged with the conspicuity maps, which produced a new image. The structure of the learner block including an example of the created conspicuity map is shown in Fig. 5.

In the process of training the YOLOR algorithm with the provided dataset, the input images are resized to $1280\times1280\times3$. These resized images are then fed into a network consisting of 24 convolution layers followed by two fully connected layers. Other hyperparameters are shown in detail in Table I.

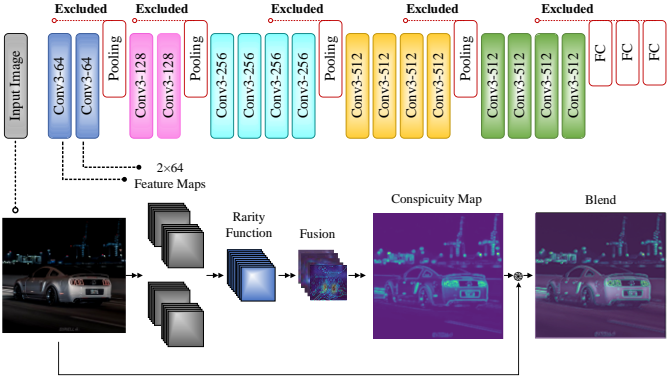

**Fig. 5.** The structure of the proposed learnable pre-processing block

TABLE I
YOLOR DETECTOR HYPERPARAMETERS

| Hyperparameters | Value | Hyperparameters | Value |
|---|---|---|---|
| Baseline Framework | YOLOv4-CSP | Initial Learning Rate | 0.0026 |
| Optimizer | Adam | Final Learning Rate | 0.2 |
| momentum | 0.949 | Pre-trained Dataset | MSCOCO |
| Batch size | 4 | Number of epochs | 110 |
| Warm-up epoch | 8 | Decay | 0.005 |

### C. Mask-Guided Attention

The ROIs obtained from the previous section include the locations of taillights and brake lights. However, there are other areas within these ROIs where the information does not have a substantial effect on the model's performance. In this section, a method inspired by the soft-attention mechanism is presented. This approach directs the network to prioritize predefined areas, optimizing the utilization of computational resources and enhancing the model's classification accuracy [32]. Based on the principles of vehicle body design, it has been conventionally observed that taillights are positioned at the utmost left and right corners of cars, positioned at a minimum height of 30 inches from the ground. Similarly, brake lights are typically situated either above or beneath the rear window or on the trunk at the same level as the taillights. In all vehicles around the world, regardless of which company they belong to, this principle remains. Based on this, a weighted binary mask for vehicles is designed with any body style (i.e., sedan, truck, SUV, MPV, etc.) and fused with the input ROIs. During the model training phase, a mask is employed to exclusively guide the network's attention towards specific pixel locations, enabling the extraction of dominant features from those regions. The focus of the model on the ROI images is not uniformly distributed and is determined based on predetermined weights. Pixels close to the left and right taillights, as well as potential brake light positions, are assigned higher weights compared to other pixels. The equation for fusing the ROIs with the mask is given in (1).

$$I_{out} = M_\omega \times I_{inp} \qquad (1)$$

Where $I_{inp}$ is the input ROI image, $M_\omega$ is the weighted binary mask, and $I_{out}$ is the output model.

The details of the design and weights of this mask, along with examples of its fusion with the ROIs obtained from the previous step, are shown in Fig. 6. The weights and dimensions reported in the design of this mask are obtained experimentally after hours of analysis. Its performance has been tested on different platforms of Ford Motor Company vehicles. Other notable characteristics of this weighted mask include i) its compatibility with vehicles from various manufacturers, allowing it to be applied universally. ii) The weighted mask effectively mitigates the impact of external disturbances, such as light reflection or rear fog lights, thereby minimizing the occurrence of false negatives and false positives in the model.

### D. Feature extraction

Deriving deep features from input data forms a fundamental aspect of deep learning techniques, which greatly influences the model's performance [33]. CNNs are a clear example of this claim [34]. In the proposed method, features extracted from the rear ROIs of the vehicle used VGGNet as one of the most common deep convolutional network architectures in the object recognition task. The design of this network started by placing a series of two convolutional layers with 64 filters 3×3. With the aim of sampling and reducing the dimension of the resulting features, a 2×2 max pooling layer with a stride of 2 is placed.

In the following, there are two convolutional layers comprising 128 filters and three convolutional layers consisting of 256 filters, with a size of 3×3. At the end of each, a 2×2 max pooling layer with a stride of 2 is embedded. Three convolutional layers with 512 filters 3×3 and a max pooling layer are the continuations of this network which are duplicated. The design of this network ended with two fully connected layers including 4096 neurons, followed by a fully connected layer of 1000 dimensions. The last convolution layer's output was linked to the network's final output. Softmax facilitated the classification prediction and generated the respective scores for each image.

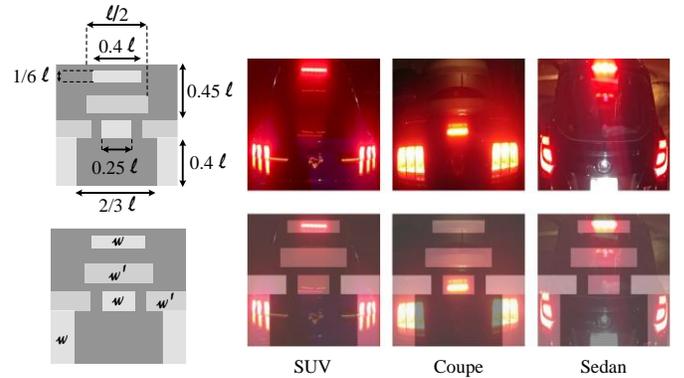

**Fig. 6.** Dimensions and weights of the designed mask along with an example of fusing with sedan, coupe, and SUV body styles. The weights of ω and ω' in this mask are 0.12 and 0.14, respectively.

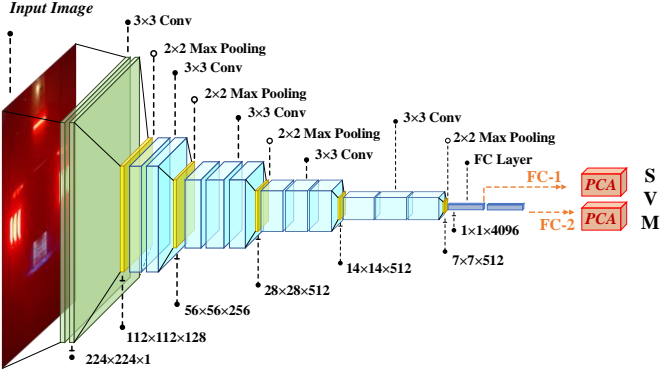

**Fig. 7.** Architecture structure designed for feature extraction, feature selection, and classification

| Hyperparameters | Value | Hyperparameters | Value |
|---|---|---|---|
| Framework | PyTorch | Learning rate | 0.001 |
| Number of parameters | 138 million | Pre-trained data | ImageNet |
| Learning rate decay interval | 30 | Optimizer | Adam |
| Batch size | 8 | LR Scheduler | StepLR |
| Number of epochs | 100 | Regularization | Dropout |

During the training step, the Softmax activation and the last fully connected layer are excluded from the network. Instead, the remaining architecture (identical to FC-2) is fed into PCA as input. The network experienced a revision in which the Softmax layer, as well as the last two fully connected layers, were eliminated. Afterwards, the network undergoes PCA for dimension reduction. Finally, the reduced dimensions are classified using a SVM classifier. The details of these steps are shown in Fig. 7.

### E. Feature selection

The last fully connected layer in the VGG-16 architecture has 4096 dimensions, which is a significant number. Working with this amount of data increases the classification error, which is called the curse of dimensionality. To overcome the curse of dimensionality, various techniques have been presented. Dimensional reduction with the help of PCA and Linear Discriminant Analysis (LDA) can be considered the most common. Furthermore, the variables within the interconnected multi-state space can be converted into a collection of independent components using a linear transformation. In such a way, with the least mean square error, the preserved information and information loss reach the minimum possible level. In this section, utilizing Principal Component Analysis (PCA), the data acquired from the last fully connected layer is transformed from a high-dimensional space to a low-dimensional space. This approach, considered as one of the feature selection methods, facilitates a more convenient evaluation of the features. Among the 4096 dimensions of data, a subset of 250 components is chosen and employed as the foundation for classification.

### F. Multi-class classification

Classification and prediction of driver behavior in four defined classes are done with the help of an SVM classifier. SVM is a type of pattern recognition method used for supervised learning in tasks such as classification and regression, first introduced in the late nineties through Vapnik's statistical learning theory [13].

Relatively simple training and not getting stuck in local maxima have made SVM perform better than CNNs for medium-dimension data. The basis of this classification is to find an optimal decision boundary so that the classes have the greatest possible distance from this boundary. The dimension of this boundary, which is called a hyperplane, depends on the meaningful features extracted from the data. Assuming the classification of $k$ classes, finding the optimal solution during training is obtained as (2):

$$\min \frac{1}{2} \sum_{i=1}^{k} w_i \times w_i + \frac{c}{n} \sum_{i=1}^{n} \xi_i \qquad (2)$$

where $w$ is the hyperplane relation, $c$ is the regularization parameter. The image pair and its label in the train-set are defined as (3):

$$S = \{(x_1, y_1), ..., (x_n, y_n)\} \qquad (3)$$

In the relation (3), for all y:

$$[x_m \cdot w_{y_m}] \geq [x_m \cdot w_y] + 100 \times \Delta(y_m, y) - \xi_m]_{m=1}^{m=n} \qquad (4)$$

where, $\Delta(y_m, y)$ is a loss function that returns zero if $y_m$ is equal to $y$, and one otherwise.

## V. DISCUSSION AND EXPERIMENTAL RESULTS

The initial raw data, along with the results acquired from the image processing methods discussed in section IV-A, are utilized as inputs for the YOLOR-based detector. This detector is responsible for identifying and extracting the Regions of Interest (ROIs) containing taillights and brake lights. These images are then resized to 224×224 and considered as input of VGGNet architecture.

At the end of this architecture, there is a layer with 4096 neurons, which are reduced to 250 dimensions with the help of PCA. Among the total data, 60% is used for training, 15% for validation and the rest for testing. The designed network is trained up to 100 epochs, and the Adam optimizer is utilized for faster convergence. Other details of network training and hyperparameters are shown in Table II. The image processing section is coded in MATLAB version 2018b software and deep learning methods with Python software with CUDA 10.0 computing platform and cuDNN 7.5.0 library. The specifications of the operating system are as follows: Windows 10 with 8 RAM, NVIDIA GeForce RTX 2060 Super 8Gb Gaming graphics card, and Intel Core i3-8100 CPU at 3.60 GHz.

TABLE III
Performance Evaluation of FC-1 and FC-2 Features in Each Class According to Different Criteria

| FC-1/FC-2 | Evaluation Criteria (%) | | | | | | | | | |
| Class | TP | TN | FP | FN | Acc. | Pre. | Spe. | Sen. | F1-Score | Kappa |
|---|---|---|---|---|---|---|---|---|---|---|
| Brake | 780 767 | 2308 2301 | 65 72 | 41 54 | 96.68 96.05 | 92.31 91.42 | 97.26 96.97 | 95.01 93.42 | 93.64 92.41 | 0.490 0.494 |
| Running | 762 773 | 2315 2299 | 66 82 | 51 40 | 96.34 96.18 | 92.03 90.41 | 97.23 96.56 | 93.73 95.08 | 92.87 92.69 | 0.499 0.491 |
| Left Turn | 693 709 | 2348 2358 | 70 60 | 83 67 | 95.21 96.02 | 90.83 92.20 | 97.10 97.52 | 89.30 91.37 | 90.06 91.78 | 0.533 0.529 |
| Right Turn | 708 674 | 2360 2353 | 50 57 | 76 110 | 96.05 94.77 | 93.40 92.20 | 97.92 97.63 | 90.31 85.97 | 91.83 88.98 | 0.529 0.541 |
| Overall | 2943 2923 | 9331 9311 | 251 271 | 251 271 | 92.14 91.52 | 92.14 91.56 | 97.38 97.17 | 92.09 91.46 | 92.10 91.46 | 0.895 0.887 |

*The green blocks indicate the best values

### A. Performance evaluation

The validation of the presented algorithm with different criteria is examined in this section. In addition to common criteria in classification problems such as accuracy, precision, and recall, other criteria are defined, and the values of each are calculated for the proposed method. The accuracy criterion, which is widely used and considered the fundamental criterion, measured the proportion of correctly predicted instances out of all the existing predictions. Although accuracy is a crucial metric for assessing model performance, it may not offer a comprehensive assessment as it fails to consider the unequal distribution of classes in the dataset. The precision criterion in binary classifications indicates the number of examples that the class algorithm correctly predicted and classified in the positive category. This criterion for multi-class classification is defined for each class. In other words, when the prediction ability of the model in a particular class is examined, this criterion would be the most appropriate solution.

Sensitivity or recall is the ability to correctly predict samples from a class, especially among all members of that class. Sensitivity and precision are two essential criteria in dealing with unbalanced classes. The specificity criterion screening classes to detect true negatives. The relationships and how to calculate each criterion are stated in (5) to (9).

$$Acc. = \frac{TP + TN}{TP + TN + FP + FN} \quad (5)$$

$$Pre. = \frac{TP}{TP + FP} \quad (6)$$

$$Spe. = \frac{TN}{TN + FP} \quad (7)$$

$$Sen. = \frac{TP}{TP + FN} \quad (8)$$

$$F1 - Score = 2 \cdot \frac{Sen. \times Pre.}{Sen. + Pre.} \quad (9)$$

Variables Acc, Pre, Spe, Sen, and Kappa are abbreviations for accuracy, precision, specificity, sensitivity, and Cohen's kappa statistic, respectively. TN, TP, FN, and FP are respectively the number of true negatives, true positives, false negatives, and false positives. The Kappa criterion is obtained as (10).

$$Kappa = \frac{p_o - p_e}{1 - p_e} \quad (10)$$

where, $p_o$ and $p_e$ are the observed probability and the expected probability, respectively, and are equal to:

$$p_o = \frac{TP + FN}{TP + TN + FP + FN} \quad (11)$$

$$p_e = \frac{TN + FP}{TP + TN + FP + FN} \quad (12)$$

In addition, in the literature, another method of displaying the performance of the model is by creating a confusion matrix. In Fig. 8, this matrix is drawn for the proposed model, where the symbols R, B, RT, and LT represent Running, Braking, Right-Turn, and Left-Turn respectively. As mentioned earlier, in the designed network, the features obtained from the second fully connected layer (FC-2) are used as the basis for classification. Again, the first fully connected layer (FC-1) is used as the basis for classification.

The results for each class of both modes are given in Table III. According to this table, using features of the first fully connected layer (FC-1) in classification shows a better performance. SVM classification is able to achieve average accuracy, precision, specificity, sensitivity, F1-score, and kappa equal to 92.14%, 92.14%, 97.38%, 92.09%, 92.10%, and 0.895, respectively. The FC-1 features are reduced to 250 nodes by PCA. However, these values for FC-2 features under identical conditions are equal to 91.52%, 91.56%, 97.17%, 91.46%, 91.46%, and 0.887, respectively.

**Fig. 8.** Confusion matrix representation for a) FC-1 and b) FC-2 model

## TABLE IV
### Showing the Performance Difference of VGGNet Architecture with FC-1 and FC-2 Features

| FC-1 / FC-2 | Evaluation Improvement (%) | | | | | | | | | | | |
|---|---|---|---|---|---|---|---|---|---|---|---|---|
| Classes | Acc. | | Pre. | | Spe. | | Sen. | | F1-Score | | Kappa | |
| # 1 | 2.10 | 1.47 | 3.92 | 3.03 | 1.39 | 1.10 | 4.15 | 2.56 | 4.03 | 2.80 | 0.7 | 0.3 |
| # 2 | 1.66 | 1.50 | 5.12 | 3.50 | 2.02 | 1.35 | 0.62 | 1.97 | 2.97 | 2.79 | 0.8 | 0.0 |
| # 3 | 0.94 | 1.75 | 0.18 | 1.19 | 0.21 | 0.21 | 4.51 | 6.58 | 2.27 | 3.99 | 1.4 | 1.8 |
| # 4 | 1.12 | 0.16 | 2.26 | 1.06 | 0.70 | 0.41 | 2.43 | 1.91 | 0.50 | 2.35 | 0.5 | 0.7 |
| Overall | *2.91* | *2.29* | *2.78* | *2.20* | *0.98* | *0.77* | *2.93* | *2.30* | *2.90* | *2.26* | *3.9* | *3.1* |

*The green blocks indicate the rate drop (negative values)

## TABLE V
### Calculation of Validation Criteria without Including the Climate Simulator Filter

| Model | Robustness Improvement (%) | | | | | |
|---|---|---|---|---|---|---|
| | Acc. | Pre. | Spe. | Sen. | F1-Score | Kappa |
| *FC-1 (~ Corruption)* | 66.56 | 66.69 | 88.84 | 66.48 | 66.48 | 0.554 |
| Robustness Drop | 25.58 | 25.45 | 8.54 | 25.61 | 25.62 | 0.341 |
| *FC-2 (~ Corruption)* | 65.25 | 65.45 | 88.41 | 65.20 | 65.19 | 0.536 |
| Robustness Drop | 26.27 | 26.11 | 8.76 | 26.26 | 26.27 | 0.351 |

### B. Ablation study

In this section, an ablative study is performed to investigate the effect of two parameters of dimension reduction and image corruption filter on the model's performance.

#### 1) Effect of dimension reduction:

In the next phase of the experiment, the VGG-16 standard architecture is substituted with feature extraction and feature selection blocks, and the training steps are repeated. This network is able to attain an accuracy of 89.23%, precision of 89.36%, specificity of 96.40%, sensitivity of 89.16%, F-1 score of 89.20%, and Kappa of 0.856. The performance difference of this architecture with the proposed FC-1 and FC-2 models for all criteria in each class is calculated in Table IV. For example, the output from the dimensionality reduction features of FC-1 and FC-2 has a higher accuracy rate of 2.91% and 2.29%, respectively, compared to VGGNet. The results indicate that the VGG-16 standard architecture obtained a much lower performance compared to other methods. This issue justifies the need to reduce the dimension of features using PCA and classification with SVM in this study.

#### 2) Effect of CNN-SVM Classifier:

Studies related to classification tasks in the literature indicate that a majority of deep learning-based methods employ Softmax activation functions to minimize their loss function. To justify the use of the SVM method in the classification process of this paper, the Softmax activation function is replaced by the SVM in the last layer. Then, all the steps are repeated in a completely similar way without the slightest change.

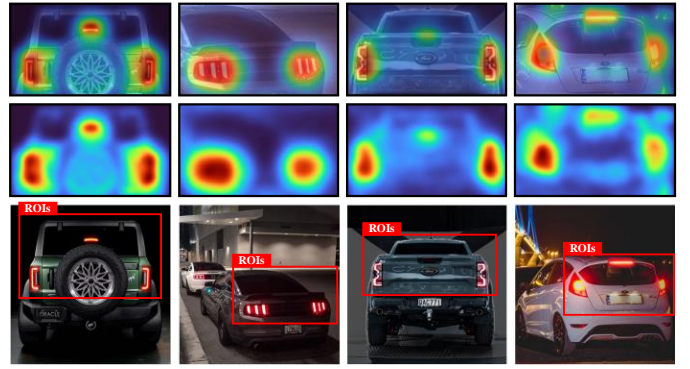

**Fig. 9.** Determining discriminative regions in image classification from visual explanation heat-maps

The results show that CNN-Softmax achieved an overall accuracy of 90.09% in multi-class classification. Meanwhile, in the previous results, a much higher accuracy is observed than the proposed CNN-SVM classification. The superior performance of CNN-SVM compared to CNN-Softmax in the problem indicates that the selection of the classifier is not only dependent on the dataset but also on an application-based approach. Therefore, a fixed rule cannot be imposed on it. For example, the authors in [35] claimed that SVMs on CIFAR10 and MNIST datasets showed a better prediction rate compared to the Softmax activation function.

#### 3) Effect of image corruption block:

Applying image corruption filters to simulate snowy, foggy, and rainy weather is another innovation of this article to improve model robustness. To check the effect of this filter on the improvement of resistance, the images made by this filter from the train-set are removed and the model is trained again. Table V displays the results attained on the test data and the reduction in model performance. As it is apparent, the design of this filter can improve the criteria defined for the performance of the model by 25%.

#### 4) Effect of mask-guided attention:

The impact of combining the mask and the original input images is demonstrated using Gradient-weighted Class Activation Mapping (Grad-CAM). Grad-Cam is one of the gradient-based algorithms that follows Class Activation Maps to visualize the effective areas in predicting the output class. This algorithm's purpose is to determine the pixels that have the greatest influence on maximizing the objective function. In this part, the masks generated from the input images are removed, and the model is retrained without them. In order to identify the influential pixels in the prediction, the gradients of the objective function are calculated concerning the outputs of the convolution layers. One of the higher-level convolution layers is visualized in Fig. 9. According to Fig. 9, Class Activation Maps focus on the left, right taillights, and brake lights, which confirm the selection of the desired mask.

TABLE VI
COMPARISON OF THE CHARACTERISTICS OF THE COLLECTED DATASET WITH OTHER SIMILAR COMMON DATASETS IN THE FIELD OF TRANSPORTATION

| Dataset | Year | # Vehicles | Make | Models | View | Insta360 | Dash-cam | Video-based | Image-based | Nighttime | Taillight Bounding Box |
|---|---|---|---|---|---|---|---|---|---|---|---|
| Barshooi [12] | 2023 | 16,244 | Ford Motor | < 2023 | Front/Rear | × | × | ✓ | × | × | ✓ |
| VLS [36] | 2022 | 7,720 | Mixed | < 2021 | Rear | × | ✓ | ✓ | × | ✓ | ✓ |
| DeepCar [27] | 2022 | 40,185 | Mixed | < 2023 | Front | × | × | ✓ | ✓ | × | × |
| Vehicle-Rear [37] | 2021 | 2,093 | Mixed | < 2020 | Rear | × | × | ✓ | × | × | × |
| H3D [38] | 2019 | 470,558 | Honda Motor | < 2018 | Rear | × | × | ✓ | ✓ | ✓ | × |
| Chen et al. [39] | 2017 | 11,452 | Mixed | < 2017 | Rear | × | ✓ | ✓ | × | ✓ | × |
| Wang et al. [40] | 2016 | 5,600 | Mixed | < 2015 | Rear | × | ✓ | ✓ | × | × | × |
| VeRi-776 [41] | 2016 | 49,357 | Mixed | < 2015 | Mixed | × | × | × | ✓ | ✓ | × |
| BMW-10 [42] | 2013 | 512 | BMW Motor | < 2013 | Mixed | × | × | × | ✓ | × | × |
| Kafai and Bhanu [43] | 2012 | 177 | Mixed | < 2011 | Rear | × | × | ✓ | ✓ | × | × |
| Ours | 2024 | 12,078 | Ford Motor | < 2024 | Rear | ✓ | ✓ | ✓ | ✓ | ✓ | ✓ |

## C. Shift domain adaptation

As mentioned earlier, the proposed patent in this work is presented to enhance the driving experience and ensure the safety of Ford Motor Company vehicles at night. For this reason, the data used in model training is collected from the products of this company. In this section, it is shown that this method can be easily applied to other vehicles without re-training the model. For this purpose, a benchmark dataset comprising 520 images of vehicles from various manufacturers such as Toyota, Hyundai, General Motors, and others is collected. Examples of these images are illustrated in Fig. 10. After that, the performance of the method is evaluated on this dataset. Although the model is not exposed to these specific vehicles during training, it exhibited accurate and highly precise class predictions. This aligns with the concept commonly known as domain adaptation.

## D. Performance comparison

This paper offers a valuable contribution from two perspectives: the collected dataset and the classification method. To the best of our knowledge, no similar datasets for the collected dataset and its annotation file can be found in the literature. However, among the existing datasets, those in the field of car classification are selected and compared with the dataset. The variety of vehicles, imaging with dash-cam and Insta360 cameras in different environmental conditions, has made it impossible to find a replacement. Additionally, the labeling of the rear lights has further added to the challenge of finding an alternative. The fact that all the images belong to one company (i.e., Ford Motor Company) makes this dataset more valuable. A comparison of proposed dataset for vehicle behavior prediction with other datasets is shown in Table VI.

TABLE VII
PERFORMANCE COMPARISON WITH OTHER STATE-OF-THE-ART METHODS ON THE COLLECTED DATASET

| Methods | Performance Comparison (%) | | | |
|---|---|---|---|---|
| | Structure | Acc. | Spe. | F1-Score |
| VGGNet | CNN | 61.18 | 87.07 | 61.07 |
| Cui et al. [44] | Hierarchical Structure | 71.63 | 90.55 | 71.73 |
| Hsu et al. [45] | CNN-LSTM | 80.71 | 95.38 | 80.69 |
| Ours | CNN-PCA-SVM | 92.14 | 97.38 | 92.10 |

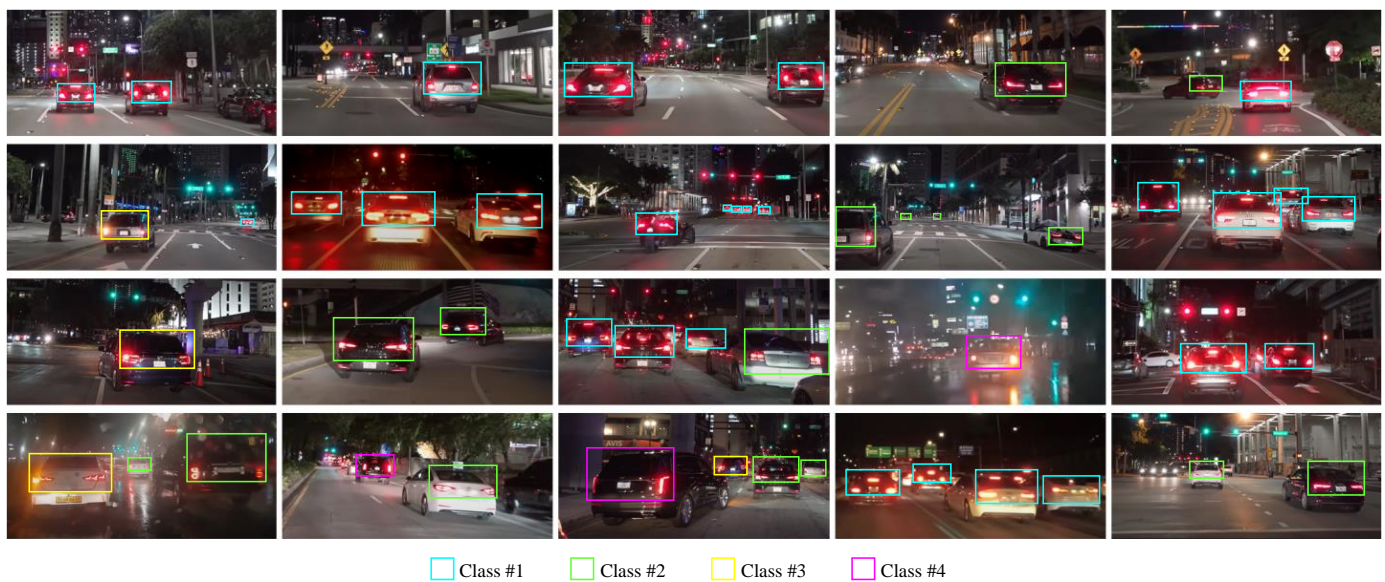

☐ Class #1  ☐ Class #2  ☐ Class #3  ☐ Class #4

**Fig. 10.** Investigating the performance of the proposed model on different types of vehicles under various environmental conditions

To prove the superiority of the proposed method over the existing methods, some of the studies are selected and re-implemented with the dataset. This causes the comparison to be made in completely similar and fair conditions. The results are reported in Table VII. The reason for the superiority of the proposed system can be examined from three aspects. First, by removing corrupted data from the train-set, the accuracy of the detectors to determine taillights or brake lights decreases dramatically. This causes the position of the taillights in some images is not determined correctly and the prediction fails. Secondly, incorporating data that simulate adverse weather conditions during the training process changes the domain when confronted with new and unseen data. Thirdly, corrupt data is considered a kind of data augmentation, which usually tries to improve performance and reduce overfitting by increasing the volume of data.

## VI. Conclusion and Future Work

In driving, every second and every reaction has great value in preventing collisions and accidents. This issue becomes more important in challenging times such as night hours and adverse weather. According to statistics, most road accidents happen at night in comparison to daytime. Therefore, drivers should always keep their focus on the brake or indicator lights of the vehicles in front of them and avoid taking their eyes off these lights.

Ford Motor Company's involvement in the advancement of autonomous vehicles, has been recently able to design a vehicle that can drive in complete darkness. To improve the performance of this vehicle, a new method for prediction the behavior of oncoming vehicles in different environmental conditions and from the taillight signals is described in this paper. This method can be applied to both autonomous vehicles with different levels and human-driven vehicles. In future works, the attempt is to sync this method with the IoT and implement it on a large scale on dash-cam.

### Acknowledgment

We sincerely acknowledge Ford Research and Innovation Center, which made this publication a reality.

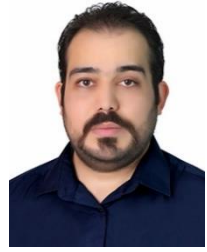

**Amir Hossein Barshooi** received the B.S. degree in electronic engineering from Shahid Beheshti University, Tehran, Iran, in 2018, and the M.S. degree in digital electronics from IUST, Tehran, Iran, in 2022. His research interests include computer vision, machine learning, deep learning, image processing, and multi-agent systems, with applications in intelligent transportation systems and self-driving cars.

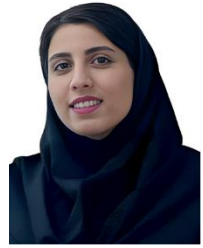

**Elmira Bagheri** received the B.S. degree in mechanical engineering from Yasouj University, Yasouj, Kohgiluyeh and Boyer-Ahmad Province, Iran, in 2018, and the M.S. degree in automotive engineering from IUST, Tehran, Iran, in 2022. Her research interests include transportation system, computer vision, deep learning, image processing, autonomous vehicles, and dynamical system.